\pdfoutput=1

\documentclass[11pt]{article}

\usepackage{emnlp2022}

\usepackage{times}
\usepackage{latexsym}

\usepackage[T1]{fontenc}

\usepackage[utf8]{inputenc}

\usepackage{microtype}

\usepackage{color}
\usepackage{graphicx}
\usepackage{paralist}
\usepackage{amsmath}
\usepackage{cleveref}
\usepackage{amssymb}
\usepackage{pifont}
\usepackage{subcaption}

\newcommand{\cmark}{\ding{51}}%
\newcommand{\xmark}{\ding{55}}

\title{EUR-Lex-Sum: A Multi- and Cross-lingual Dataset for Long-form Summarization in the Legal Domain}

\author{Dennis Aumiller\thanks{~~These authors contributed equally to this work.}${~\,^\dagger}$, Ashish Chouhan$^{*\dagger\ddag}$ \and Michael Gertz$^\dagger$ \\
  $^\dagger$ Institute of Computer Science, Heidelberg University \\
  $^\ddag$ School of Information, Media and Design, SRH Hochschule Heidelberg\\
  \texttt{\{aumiller, chouhan, gertz\}@informatik.uni-heidelberg.de}\\}

\begin{document}
\maketitle
\begin{abstract}
Existing summarization datasets come with two main drawbacks: 
(1) They tend to focus on overly exposed domains, such as news articles or wiki-like texts, and 
(2) are primarily monolingual, with few multilingual datasets.
In this work, we propose a novel dataset, called EUR-Lex-Sum, based on manually curated document summaries of legal acts from the European Union law platform (EUR-Lex).
Documents and their respective summaries exist as cross-lingual paragraph-aligned data in several of the 24 official European languages, enabling access to various cross-lingual and lower-resourced summarization setups. 
We obtain up to 1{,}500 document/summary pairs per language, including a subset of 375 cross-lingually aligned legal acts with texts available in \emph{all} 24 languages.\\
In this work, the data acquisition process is detailed and key characteristics of the resource are compared to existing summarization resources.
In particular, we illustrate challenging sub-problems and open questions on the dataset that could help the facilitation of future research in the direction of domain-specific cross-lingual summarization.
Limited by the extreme length and language diversity of samples, we further conduct experiments with suitable extractive monolingual and cross-lingual baselines for future work.\\
Code for the extraction as well as access to our data and baselines is available online at: \url{https://github.com/achouhan93/eur-lex-sum}. 
	


\end{abstract}

\section{Introduction}
Despite a long history in the field of text summarization~\cite{luhn-1958-automatic}, current systems in the area are still mainly targeted towards a few select domains.
This stems in part from the homogeneity of existing summarization datasets and extraction processes: frequently, these are either collected from news articles~\cite{duc-2004, sandhaus-2008-new, DBLP:conf/nips/HermannKGEKSB15, narayan-etal-2018-dont, grusky-etal-2018-newsroom,  hasan-etal-2021-xl} or wiki-style knowledge bases~\cite{ladhak-etal-2020-wikilingua, frefel-2020-summarization}, where alignment with supposed ``summaries'' is particularly straightforward.
Domain outliers do exist, e.g., for scientific  literature~\cite{cachola-etal-2020-tldr} or the legal domain~\cite{mediatum, kornilova-eidelman-2019-billsum, manor-li-2019-plain, DBLP:conf/ecir/BhattacharyaHRP19}, but are primarily restricted to the English language or do not contain finer-grained alignments between cross-lingual documents.\\
Reasons for the usage of mentioned predominant domains are manifold: Data is reasonably accessible throughout the internet, can be automatically extracted, and the structure naturally lends itself to the extraction of excerpts that can be seen as a form of summarization.
For news articles, short snippets (or headlines) describing the gist of main article texts are quite common. Wikipedia has an introductionary paragraph that has been framed as a ``summary'' of the remaining article~\cite{frefel-2020-summarization}, whereas others utilize scholarly abstracts (or variants thereof) as extreme summaries of academic texts~\cite{cachola-etal-2020-tldr}.\\
For a variety of reasons, using these datasets as a training resource for summarization systems introduces (unwanted) biases. Examples include extreme lead bias~\cite{zhu-etal-2021-leveraging}, 
focus on extremely short input/output texts~\cite{narayan-etal-2018-dont}, or high overlap in the document contents~\cite{nallapati-etal-2016-abstractive}. 
Models trained in such a fashion also tend to score quite well on zero-shot evaluation of datasets from similar domains, however, poorly generalize beyond immediate in-domain samples that follow a different content distribution or longer expected summary length.\\
Simultaneously, high-quality multilingual and cross-lingual data for training summarization systems is scarce, particularly for datasets including more than two languages.
Existing resources are often constructed in similar fashion to their monolingual counterparts~\cite{scialom-etal-2020-mlsum, varab-schluter-2021-massivesumm} and subsequently share the same shortcomings of low data quality.\\
Our main contribution in this work is the construction of a novel multi- and cross-lingual corpus of reference texts and human-written summaries that extract texts from legal acts of the European Union (EU).
Aside from a varying number of training samples per language, we provide a paragraph-aligned validation and test set across all 24 official languages of the European Uninon\footnote{\url{https://eur-lex.europa.eu/content/help/eurlex-content/linguistic-coverage.html}, last accessed: 2022-06-15}, which further enables cross-lingual evaluation settings.


\section{Related Work}
Influencing works can generally be categorized into works about EU data, or more broadly about summarization in the legal domain. Aside from that, we also compare our research to other existing multi- and cross-lingual works for text summarization.

\subsection{The EU as a Data Source}
Data generated by the European Union has been utilized extensively in other sub-fields of Natural Language Processing.
The most prominent example is probably the Europarl corpus~\cite{koehn-2005-europarl}, consisting of sentence-aligned translated texts generated from transcripts of the European Parliament proceedings, frequently used in Machine Translation systems due to its size and language coverage.\\
In similar fashion to parliament transcripts, the European Union has its dedicated web platform for legal acts, case law and treaties, called EUR-Lex~\cite{bernet-berteloot-2006-eur}\footnote{most recent URL: \url{https://eur-lex.europa.eu}, last accessed: 2022-06-15}, which we will refer to as the \emph{EUR-Lex platform}.
Data from the EUR-Lex platform has previously been utilized as a resource for extreme multi-label classification~\cite{mencia-furnkranz-2010-efficient}, most recently including an updated version by \citet{chalkidis-etal-2019-extreme, chalkidis-etal-2019-large}. In particular, the MultiEURLEX dataset~\cite{chalkidis-etal-2021-multieurlex} extends the monolingual resource to a multilingual one, however, does not move beyond the classification of EuroVoc labels.
To our knowledge, document summaries of legal acts from the platform have recently been used as a monolingual English training resource for summarization systems~\cite{klaus-etal-2022-summarizing}.

\subsection{Processing of Long Legal Texts}
Recently, using sparse attention, transformer-based models have been proposed to handle longer documents~\cite{beltagy2020longformer, zaheer-etal-2020-big}. However, the content structure is not explicitly considered in current models. \citet{Yang2020} proposed a hierarchical Transformer model, SMITH, that incrementally encodes increasingly larger text blocks. 
Given the lengthy nature of legal texts, \citep{aumiller-etal-2021-structural} investigate methods to separate content into topically coherent segments, which can benefit the processing of unstructured and heterogeneous documents in long-form processing settings with limited context.
From a data perspective, \citet{kornilova-eidelman-2019-billsum} propose BillSum, a resource based on US and California bill texts, spanning between approximately 5{,}000 to 20{,}000 characters in length.
For the aforementioned English summarization corpus based on the EUR-Lex platform, \citet{klaus-etal-2022-summarizing} utilize an automatically aligned text corpus for fine-tuning BERT-like Transformer models on an extractive summarization objective. Their best-performing approach is a hybrid solution that prefaces the Transformer system with a TextRank-based pre-filtering step.

\subsection{Datasets for Multi- or Cross-lingual Summarization}
\label{subsec:cross}
For Cross-lingual Summarization (XLS), \citet{wang2022survey} provide an extensive survey on the currently available methods, datasets, and prospects. Resources for XLS can be divided into two primary categories: synthetic datasets and web-native multilingual resources. For the former, samples are created by directly translating summaries from a given source language to a separate target language. Examples include English-Chinese (and vice versa) by~\citet{zhu-etal-2019-ncls}, and an English-German resource~\cite{bai-etal-2021-cross}. Both works utilize news articles for data and neural MT systems for the translation. In contrast, there also exist web-native multilingual datasets, where both references and summaries were obtained primarily from parallel website data. Global Voices~\cite{nguyen-daume-iii-2019-global}, XWikis~\cite{perez-beltrachini-lapata-2021-models}, Spektrum~\cite{fatima-strube-2021-novel}, and CLIDSUM~\cite{wang2022clidsum} represent instances of datasets for the news, encyclopedic, and dialogue domain, with differing numbers of supported languages.

\noindent We have previously mentioned some of the multilingual summarization resource where multiple languages are covered.
MLSUM~\cite{scialom-etal-2020-mlsum} is based on news articles in six languages, however, without cross-lingual alignments. Similarly without alignments, but larger in scale, is MassiveSum~\cite{varab-schluter-2021-massivesumm}.
XL-Sum~\citet{hasan-etal-2021-xl} does provide document-aligned news article, in 44 distinct languages, extracted data from translated articles published by the BBC. In particular, their work also provides translations in several lower-resourced Asian languages.
WikiLingua~\cite{ladhak-etal-2020-wikilingua} borders the multi- and cross-lingual domain; some weak alignments exist, but only for English references, and not between languages themselves.


\section{The EUR-Lex-Sum Dataset}
We present a novel dataset based on available multilingual document summaries from the EUR-Lex platform. The final dataset, which we title ``\emph{EUR-Lex-Sum}'', consists of up to 1{,}500 document/summary pairs per language. For comparable validation and test splits, we identified a subset of 375 cross-lingually aligned legal acts that are available in all 24 languages. 
In this section, the data acquisition process is detailed, followed by a brief exploratory analysis of the documents and their content. Finally, key intrinsic characteristics of the resource are compared with relation to existing summarization resources.
In short, we find that the combination of human-written summaries coupled with comparatively long source \emph{and} summary texts makes this dataset a suitable resource for evaluating a less common summarization setting, especially for long-form tasks.

\subsection{Dataset Creation}
\label{sec:acq}
The EUR-Lex platform provides access to various legal documents published by organs within the European Union. In particular, we focus on currently enforced EU legislation (legal acts) for the 20 domains from the EUR-Lex platform.\footnote{\url{https://eur-lex.europa.eu/browse/directories/legislation.html}, last accessed: 2022-06-21}
From the mentioned link, direct access to lists of published legal acts associated with a particular domain is available, which forms the starting point for our later crawling step.
Notably, each of these domains also provides a diverse set of specific keywords, topics and regulations, which even within the dataset provide a high level of diversity.\\
A legal act is uniquely identified by the so-called Celex ID, composed of codes for the respective sector, year and document type. The ID is consistent across all 24 languages, which makes it possible to align articles on a document level.
Across all 20 domains, the website reports a total of 26{,}468 legal acts spanning from 1952 until 2022. However, as there is a probability of a particular legal act being assigned to multiple domains, approximately 22{,}000 unique legal acts can be extracted from the platform. We do not consider EU case law and treaties, which are also available through the EUR-Lex platform, but in other document formats.

\subsubsection{Crawling}
The web page of a particular legal act contains the following page content relevant for a summarization setting:
\begin{inparaenum}
	\item The published text of the particular legal act in various file formats,	
	\item metadata information about the legal acts, such as published year, associated treaties, etc.,
	\item links to the content pages in other official languages, and
	\item if available, a link to an associated summary document.
\end{inparaenum}\\ 
This work contributes to preparing a dataset with the legal act content and their respective summaries in different languages. Therefore, crawling over the entirety of published legal acts gives access to all relevant information needed to extract source and summary text pairs. 
Since a single legal act requires 50 individual web requests to extract files across all languages, we have a total of around 5.5 million access requests, distributed across the span of a month between May and June 2022.
We dump the content of all accessed acts in a local Elasticsearch instance, and separately mark documents without existing associated summaries.
This allows the resource to be continually updated in the future without re-crawling documents that do not have available summaries.

\subsubsection{Filtering}
\label{sec:filtering}
For further processing, we filter the documents available through our offline storage.
First, some article texts may only be available as scanned (PDF) documents, which compromises text quality and is therefore discarded.
For the most consistent representation, we choose to limit ourselves to articles present in an HTML document, with further advantages explained in \Cref{subsec:dataquality}.
Availability of HTML documents generally correlates with the publishing year, see \Cref{sec:temporal}, presumably due to the emergence of the world wide web during the 1990s.
Similarly, a document is not required to have an associated summary, limiting sample pairs' availability. A full distribution of available HTML sample pairs can be found in \Cref{fig:langs}. We could not identify any particular reasoning behind what documents do have summaries and which do not.

More problematic is the fact that between 20-30\% of the available summaries (depending on the language) are associated with \emph{several} source documents, essentially turning this into a multi-document summarization setting.
Since this work focuses exclusively on single document summarization, we pair the summary with the longest associated reference document to maximize availability.
\Cref{tab:novelties} details the impact of considering only the longest document in terms of $n$-gram novelty; we observe a consistent increase of novel $n$-grams by about 5 percentage points over the subset of single-reference documents.
While the concatenation of all relevant reference documents would eliminate any difference in $n$-gram overlap between the summary and reference texts, having a single reference document conserves the correct processing of lead biases over alternatives that aggregate several texts.
Further, concatenation leads to ambiguous text orderings, which may change summarization outcomes based on different aggregation strategies.
However, the subset of these multi-document samples could be a challenging problem based on our available corpus that may be explored in a separate context for future work.\\
Finally, we filter out all document pairs where the reference text is shorter than the input document. This occurs only for multi-document summary pairs, where sometimes several short acts are aggregated into a single summary.

\begin{table}[t]
	\setlength{\tabcolsep}{2.5pt}
	\begin{tabular}{l|cccc}
		 & \multicolumn{4}{c}{\textbf{$n$-gram novelty}} \\
		 \textbf{Subset} & $1$-gram & $2$-gram & $3$-gram & $4$-gram \\
		 \hline
		 All samples & 42.25 & 64.07 & 77.34 & 83.73 \\
		 Single-ref subset & 41.74 & 63.52 & 76.87 & 83.33 \\
		 Longest & 46.77 & 68.83 & 81.44 & 87.18 \\
		 Concatenated & 41.03 & 63.06 & 76.38 & 82.77
	\end{tabular} 
	\caption{Comparison of $n$-gram novelties for the English subset, differentiating by the number of reference documents. \emph{Longest} considers the subset of multi-reference documents with only the longest document as a reference; \emph{Concatenated} uses the concatenation of all associated references.}
	\label{tab:novelties}
\end{table}



\noindent After filtering out invalid samples, between 391 (Irish) to 1{,}505 (French) documents remain; the full list of samples broken down by language can be found in the Appendix in \Cref{tab:langs}. Across all languages, we manage to extract 31{,}987 pairs.

\subsubsection{Data Split}
To ensure a suitable (and comparable) validation and test split across different languages, all documents having sample pairs available in 24 languages (375 total) are taken out of the available respective subsets. Of the 375 documents, 187 samples are randomly selected into the validation set, and the remaining 188 are taken as the corresponding test set.
All other documents are assigned to the language-dependent training sets. No guarantee for cross-lingual availability is provided for the training set, however, most documents do appear in several of the languages.
In particular, 
We will use these filtered data splits for future experiments in this paper unless explicitly mentioned otherwise.

\section{Exploratory Analysis}
An exploratory analysis of the dataset is conducted to confirm the resource's viability for automatic summarization and overall data quality. Aside from a qualitative view of the resource and an analysis of the temporal distribution of our samples, we provide a comprehensive look at intrinsic metrics commonly used for summarization datasets.

\begin{figure}[t]
	\centering
	\includegraphics[width=0.5\textwidth]{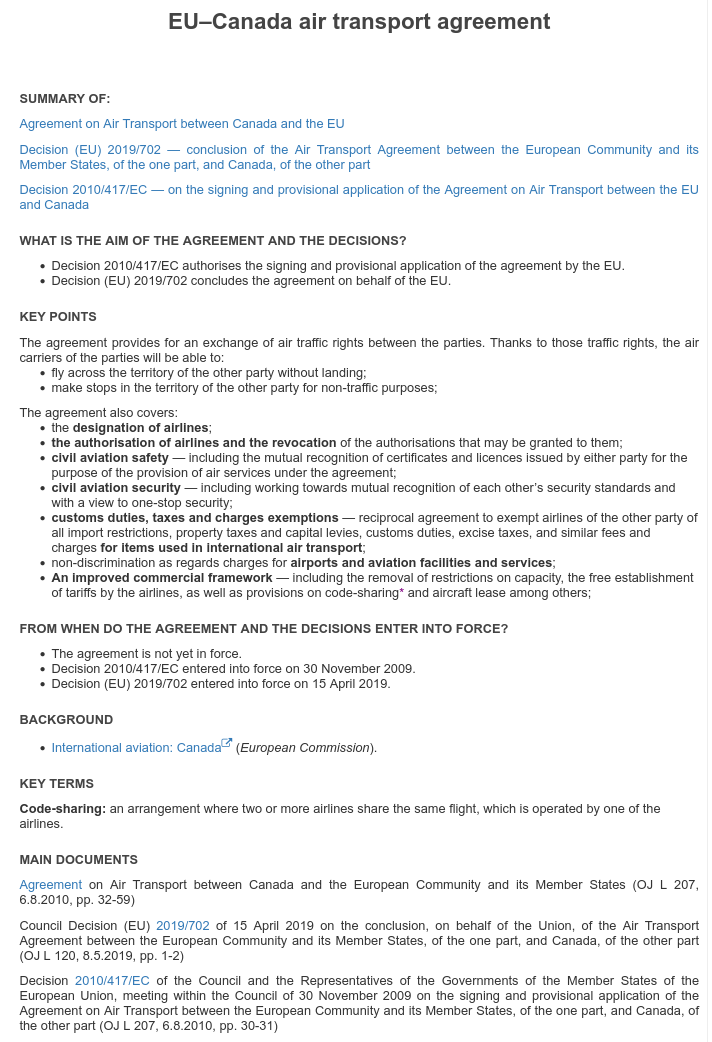}
	\caption{Summary of legal act with Celex ID 32019D0702. Visible are several distinct sections, with the majority of the document describing key points of the underlying legal acts. This particular summary aggregates content from several legal acts, of which we consider the longest one as the reference document.}
	\label{fig:outline}
\end{figure}

\subsection{Data Quality}
\label{subsec:dataquality}
Documents of the EU are generally held to a high standard, and the legal acts are no exception. This also extends to the summaries, which follow a particular set of guidelines for their creation process.\footnote{\url{https://etendering.ted.europa.eu/cft/cft-documents.html?cftId=6490}, last accessed: 2022-06-15.}
In particular, guidelines for drafting summary texts are detailed in Technical Annex I, which specify several key instructions for generating human-written summaries of an underlying legal act.
Most prominently, they recommend a target length for key point summaries between 500-700 words and formulate a template structure for the overall text outline. An example of a typical summary structure can be seen in \Cref{fig:outline}.
Aside from the key points, this includes, e.g., references to the main documents or specific act-related key phrases.
We want to highlight that the generation guidelines changed over time. Since we do not have access to previous versions of the guidelines, we manually probed comparisons between older and newer documents, which exposed a highly structural similarity despite changes in guidelines.

\noindent The published documents and summaries offer further peculiarities in both their content structure as well as the creation process:
First, the multilingual versions of both documents and summaries are always translated from the original English legal act (or English summary thereof),\footnote{This has been confirmed by the Publications Office of the European Union in private correspondence.} which ensures strict content similarity of the same text across all available languages.
Second, due to their HTML representation, it is possible to extract \emph{paragraph-aligned} texts between language-specific versions. This is a well-known property of EU-level data, most notably exploited in the Europarl corpus~\cite{koehn-2005-europarl} for automatic alignments of machine translation training data.
We similarly maintain this structure during the extraction process to use it at later stages, e.g., for more informed evaluation setups or cross-lingual pre-training.


\subsection{Temporal Distribution}
\label{sec:temporal}
\begin{figure}[t]
	\centering
	\includegraphics[width=0.5\textwidth]{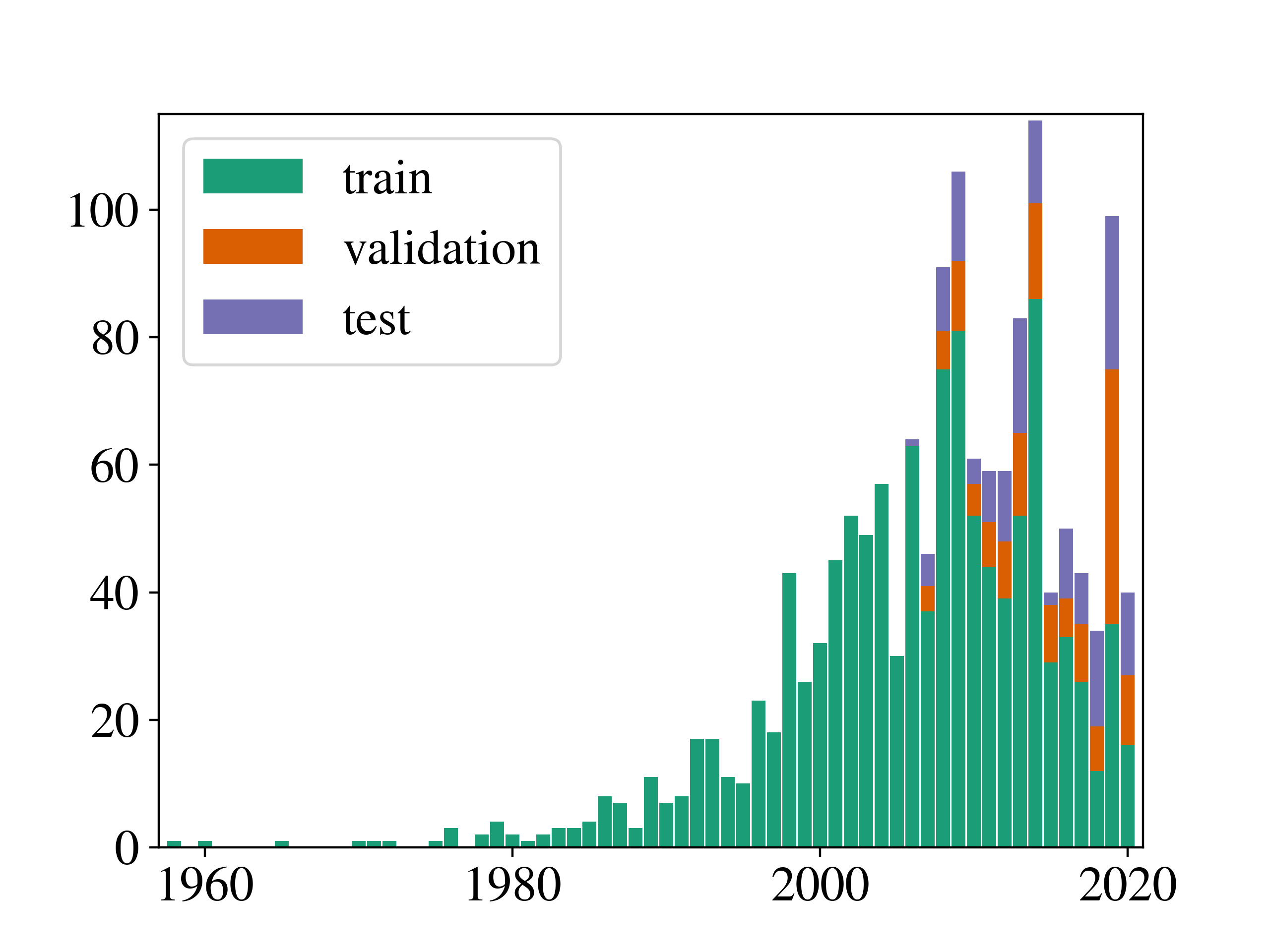}
	\caption{Distribution of the publishing year of unique legal acts included in the final dataset. Availability of documents increases after 1990.}
	\label{fig:years}
\end{figure}

\Cref{fig:years} displays the distribution of filtered documents by the year of publication. The amount of available samples increases after 1990, which likely coincides with more member states joining, as well as a shift to digital archiving (compared to OCR scans of PDF documents, which are excluded from our corpus). Compared to other European resources, such as Multi-EURLex~\cite{chalkidis-etal-2021-multieurlex}, a lesser topical shift is expected in our resource, simply due to a more limited time frame.
Notably, we also include the distribution by dataset split and observe an even stronger bias towards more recent legal acts for validation and test sets. This is a natural consequence of the requirement for validation and test sets that legal acts be present in all 24 languages, which includes more recently added official language, such as Croatian (added in 2013) or Irish (added in 2022). 
We also want to mention that amendments to both reference and summary texts might be added (or revised) several years after their original publication, which is not reflected in our analysis.

\begin{figure*}[ht]
	\centering
	\begin{subfigure}[b]{0.327\textwidth}
		\includegraphics[width=1.0\textwidth]{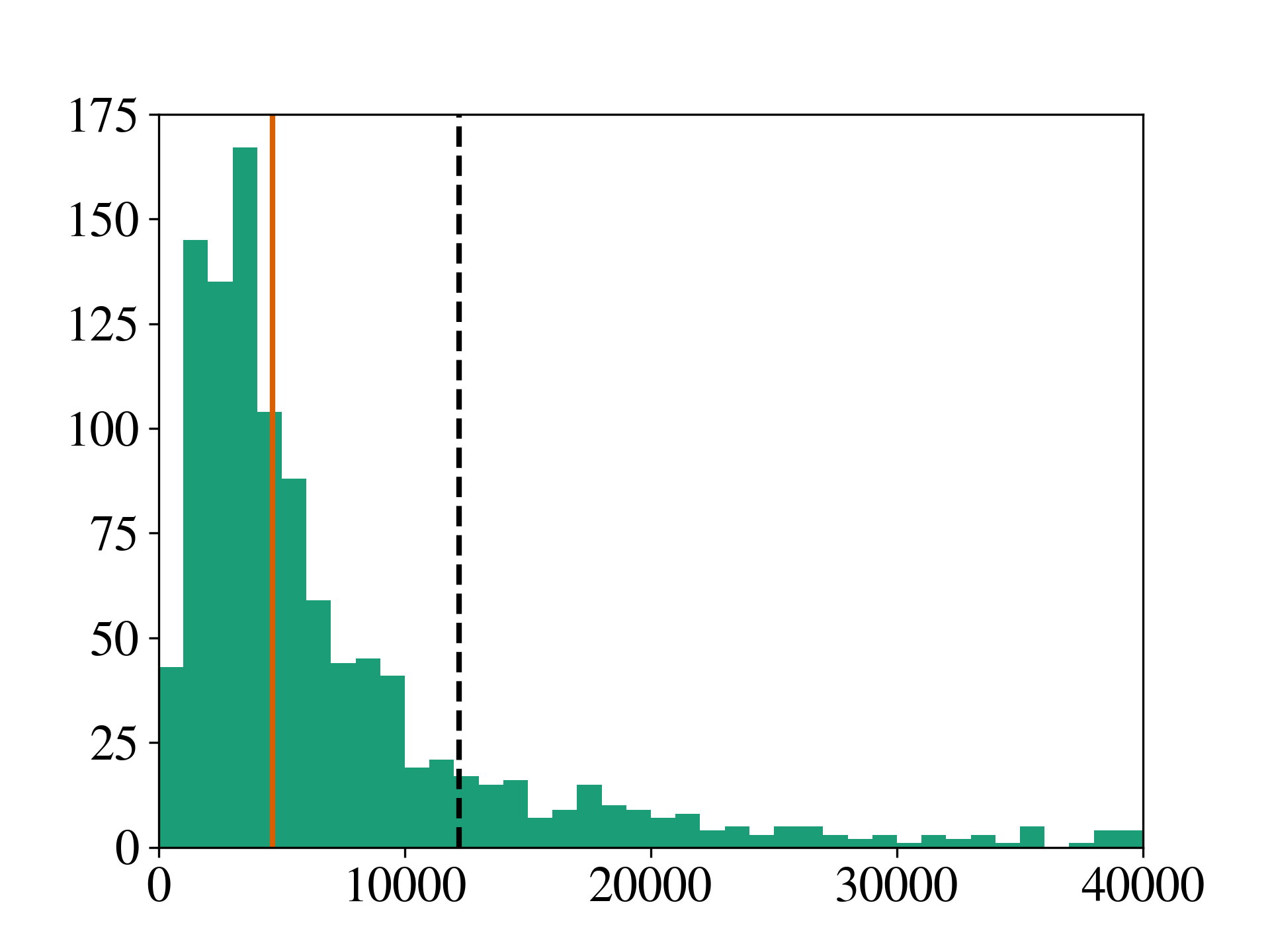}
		\caption{Reference tokens}
	\end{subfigure}
	\begin{subfigure}[b]{0.327\textwidth}
		\includegraphics[width=1.00\textwidth]{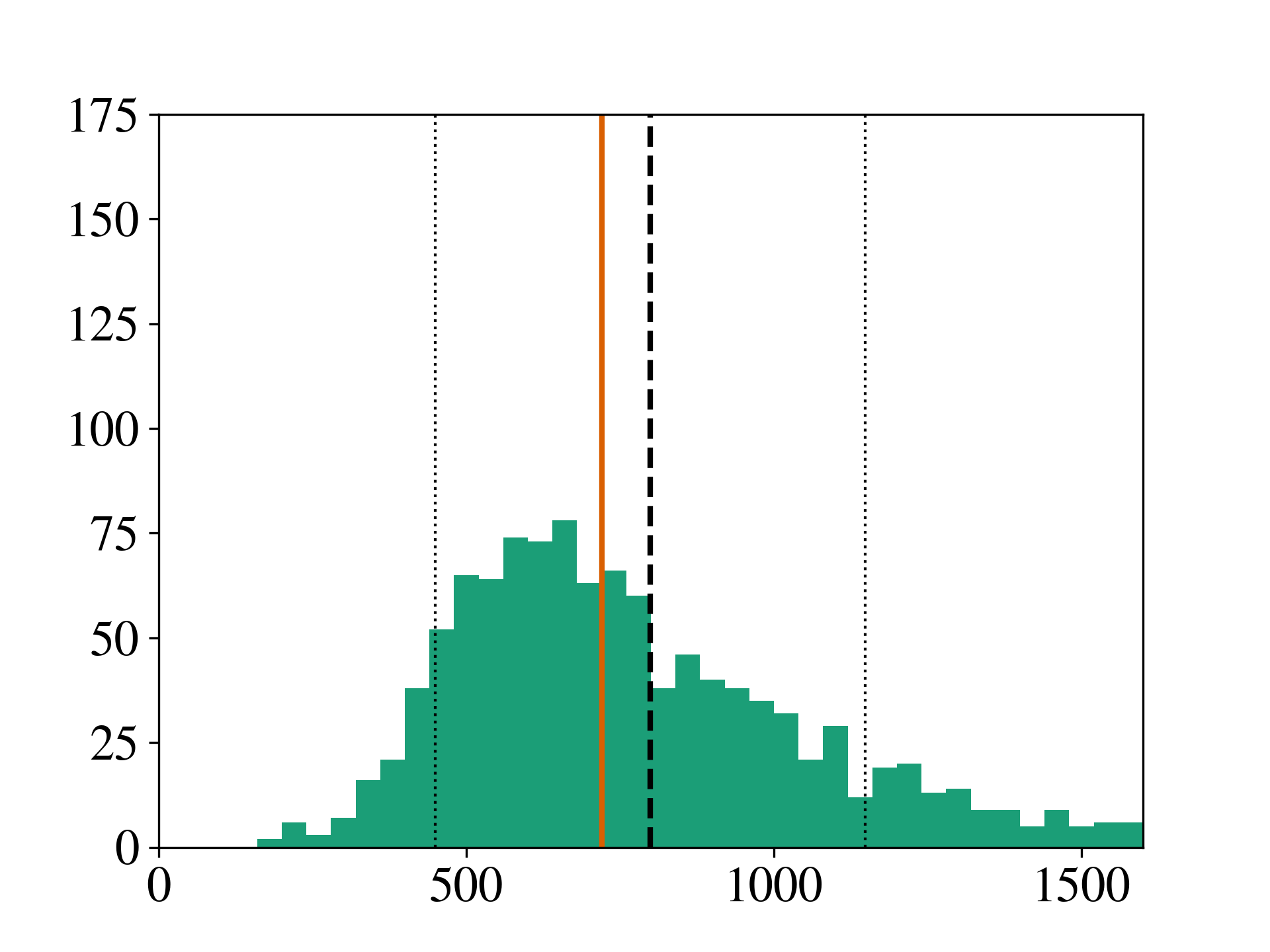}
		\caption{Summary tokens}
	\end{subfigure}
	\begin{subfigure}[b]{0.327\textwidth}
		\includegraphics[width=1.00\textwidth]{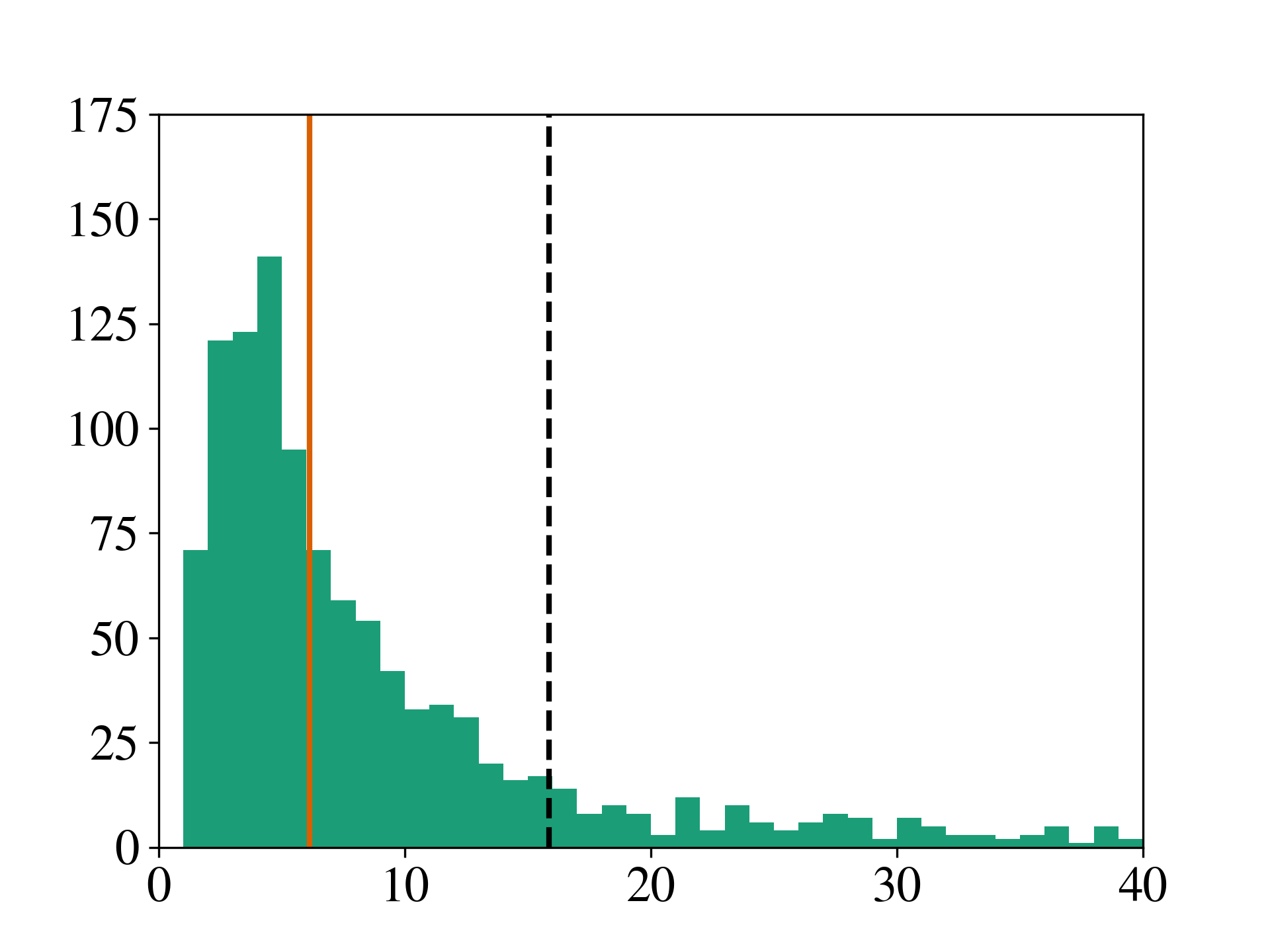}
		\caption{Token compression ratio}
	\end{subfigure}
	\caption{Histogram of the English training set, comparing article token lengths. Displayed are the distribution for references (left), summaries (center), and compression ratios (right). Vertical lines show median length (continuous orange), mean length (dashed black), and standard deviation (dotted black lines). The latter exceeds display limits for reference length and compression ratio. Ranges are limited to the 95th length percentile for legibility.}
	\label{fig:stats}
\end{figure*}

\subsection{Document Structure}

An example of the content structure of a document summary is provided in \Cref{fig:outline}. The formatted text reveals significant sections of the summary, where the majority is taken up by free text describing the key goals and highlights of the sub-points within a longer legal act. It further describes which legal act (or several acts) are associated with the summary. As previously described, we limit ourselves to the longest associated legal act for a summary associated with several acts.\\
While we provide raw text for the extracted legal act document in the proposed resource, example document in \Cref{fig:outline} reveals a potential use case of semi-structured visual information from HTML tags (e.g., headline descriptors or bullet lists), which could be used for a fine-grained distinction between different content parts.
In our preliminary experiments, we found that the used HTML tags for content elements can vary significantly between different legal acts (e.g., using modified \texttt{div} containers instead of \texttt{H3} for sub-headings) and therefore keep the inclusion of such features for future work.

\subsection{Summarization-related Dataset Metrics}
We adopt metrics from prior work to automatically analyze summarization datasets~\cite{grusky-etal-2018-newsroom, zhong-etal-2019-closer, bommasani-cardie-2020-intrinsic}. Our corpus reveals a high degree of abstractivity, which is surprising given the enormous length of input texts.

\begin{table*}
	\centering
	\begin{tabular}{l|cc|cc|c|cccc}
		& \multicolumn{2}{c|}{\textbf{Reference tokens} } &\multicolumn{2}{c|}{\textbf{Summary tokens} } & \textbf{Comp.} & \multicolumn{4}{|c}{\textbf{\% novel $n$-grams in summary}} \\
		& \ \textbf{Min} & \textbf{Max} & \ \textbf{Min} & \textbf{Max} & \textbf{Ratio} & \textbf{$1$-gram} & \textbf{$2$-gram} & \textbf{$3$-gram}& \textbf{$4$-gram} \\
		\hline
		Train &  385 & 1{,}087{,}217 & 173 & 3021 & 16 $\pm$ 62 & 44.10 & 65.97 & 78.85 & 84.96 \\
		Val. & 1{,}143 & 199{,}405 & 354 & 5136 & 18 $\pm$ 17 & 36.65 & 58.23 & 72.74 & 79.96 \\
		Test & 1{,}544 & 403{,}319 & 369 & 2987 & 18 $\pm$ 20 & 36.78 & 58.46 & 72.83 & 80.07
	\end{tabular}
	\caption{Complementary dataset properties ($\min$ and $\max$ token lengths of both reference texts and summaries), as well as compression ratio, across subsets. We further report novelty $n$-gram shares in the gold summary. Values are computed on the English subset splits.}
	\label{tab:en_stats}
\end{table*}

\subsubsection{Length Distribution}
Based on the fact mentioned in \Cref{subsec:dataquality} that documents are created as translations from the English original, we focus more on the distribution of legal acts and their summary lengths in English as a representative language. A more exhaustive overview can be found in the Appendix in \Cref{tab:langs}, which gives more insight into language-specific length variations due to document availability, or simply morphological/syntactic differences, e.g., compound words.

\noindent Histogram plots in \Cref{fig:stats} show a Zipfian distribution for reference text lengths, with a mean of around 12{,}000 tokens; however, we also observe an exceptionally large standard deviation due to extreme outliers, mentioned in \Cref{tab:en_stats}.
In contrast, summary lengths exhibit closer to a normal distribution, which matches the guideline document's suggested length of 500-700 words. The observed mean is slightly higher at around 800 tokens, which can be attributed to document overhead not counting towards the actual summarizing content, such as referenced documents and separately highlighted key concepts. However, we also observe extreme outliers for summary texts (cf.~\Cref{tab:en_stats}).

\subsubsection{Compression Ratio}
We follow the definition of an unrestricted compression ratio~\cite{grusky-etal-2018-newsroom}, dividing the (token) length of an article by the associated summary (token) length.
This carries the same semantic value as inverse definitions of compression ratio, such as used by~\citet{bommasani-cardie-2020-intrinsic}.
When looking at the token-level compression ratio displayed in \Cref{fig:stats}, a comparatively high mean is observed, despite extremely long summary documents. Comparing compression ratios reported by~\cite{zhong-etal-2019-closer} for news-based datasets indicates that EUR-Lex-Sum has a mean compression ratios similar to the CNN/DailyMail~\cite{DBLP:conf/nips/HermannKGEKSB15} and NYT~\cite{sandhaus-2008-new} corpora.

\subsubsection{$n$-gram Novelty}
To provide insight into the abstractiveness of gold summaries, we follow \citet{narayan-etal-2018-dont} in analyzing the fraction of $n$-grams not present in the original reference article. This metric is similar to content coverage metrics used by~\citet{grusky-etal-2018-newsroom} or \citet{zhong-etal-2019-closer}.
When comparing novelty $n$-grams reported in \Cref{tab:en_stats}, it should be noted that this slightly overestimates the real value. This can be attributed to our use of whitespace tokenization, which may cause more $n$-grams due to decreased tokenization accuracy; we further discuss the tokenization choice in \Cref{sec:experiments}.

\section{Experiments}
\label{sec:experiments}
As a reference for future work building on top of this dataset, we provide a set of suitable baselines and discuss limitations of methods and data. Notably, there are considerable challenges in constructing baseline runs with popular algorithms on this dataset if trying to cover all languages.\\
Primarily, even summary lengths exceed input limitations of popular abstractive neural models based on transformer architectures; these systems are generally limited to 512 (subword) tokens~\cite{lewis-etal-2020-bart, xue-etal-2021-mt5}, and even length-focused alternatives generally boast only up to 4096 tokens~\cite{beltagy2020longformer, DBLP:conf/nips/ZaheerGDAAOPRWY20}, which is well below the median length of reference texts and prevents us from training systems without further (manual) alignments provided on chunks of the input text.\\
Less obvious, but no less problematic is the availability of tokenizers or sentence splitting methods in popular NLP libraries, affecting several languages in our corpus (for a more in-depth list of supported languages by library, see Appendix \Cref{tab:langs}). This inherently prevents fair sentence-level evaluation (or extraction), as system performance is not guaranteed for underrepresented languages.

\noindent Aside from a set of extractive baselines, we further evaluate a cross-lingual scenario in which summaries for the English reference text are generated and then translated into the target languages. The hypothesis is that this provides insight into limitations of existing XLS systems discussed in \Cref{subsec:cross} and also represents more realistic deployment scenarios where XLS systems can be utilized as supportive summarizers for monolingual input texts.

\subsection{Zero-shot Extractive Baselines}

\begin{table}[t]
	\setlength{\tabcolsep}{2.5pt}
	\centering
	\hspace*{-1em}
	\begin{tabular}{l|ccc||ccc}
		&  \multicolumn{3}{c||}{\textbf{Validation} } & \multicolumn{3}{c}{\textbf{Test}} \\
		& \textbf{R-1} & \textbf{R-2} & \textbf{R-L} & \textbf{R-1} & \textbf{R-2} & \textbf{R-L} \\
		\hline
		English & 25.99 & 13.34 & 13.30 & 26.68 & 13.65 & 13.58 \\
		French & 32.18 & 18.03 & 15.15 & 32.35 & 18.00 & 15.16 \\
		German & 26.00 & 13.12 & 12.24 & 26.72 & 13.75 & 12.56 \\
		Spanish & 27.04 & 16.43 & 14.75 & 28.34 & 17.12 & 15.23 \\
		Italian & 27.29 & 14.01 & 12.63 & 28.57 & 14.24 & 12.90 \\
		Portuguese & 30.12 & 17.17 & 15.08 & 30.67 & 17.20 & 15.20 \\
		Dutch & 29.07 & 14.92 & 14.66 & 29.62 & 14.76 & 14.73\\
		Danish & 28.78 & 13.90 & 13.14 & 29.22 & 13.86 & 13.19 \\
		Greek & 24.42 & \ \ 9.77 & 15.46 & 24.79 & \ \ 9.45 & 15.46 \\
		Finnish & 26.40 & 11.88 & 11.87 & 26.49 & 11.68 & 11.80 \\
		Swedish & 30.25 & 15.40 & 14.27 & 30.67 & 15.47 & 14.35 \\
		Romanian & 35.69 & 16.08 & 14.90 & 34.75 & 15.16 & 14.59 \\
		Hungarian & 33.71 & 19.53 & 15.49 & 34.55 & 19.69 & 15.64 \\
		Czech & 30.96 & 16.65 & 14.16 & 31.86 & 16.76 & 14.32 \\
		Polish & 28.47 & 14.42 & 12.68 & 28.88 & 14.42 & 12.73 \\
		Bulgarian & 26.36 & \ \ 9.15 & 16.54 & 25.58 & \ \ 8.40 & 16.13 \\
		Latvian & 31.24 & 15.55 & 12.99 & 31.73 & 15.77 & 13.15\\
		Slovene & 26.75 & 12.25 & 11.64 & 27.19 & 12.34 & 11.79 \\
		Estonian & 26.33 & 11.64 & 11.84 & 26.39 & 11.41 & 11.66 \\
		Lithuanian & 26.79 & 12.43 & 11.44 & 26.76 & 12.45 & 11.59 \\
		Slovak & 30.30 & 15.04 & 13.14 & 30.65 & 14.94 & 13.14 \\
		Maltese & 29.71 & 14.55 & 12.73 & 30.51 & 14.62 & 12.86 \\
		Croatian & 33.50 & 13.46 & 13.50 & 32.64 & 12.76 & 13.29 \\
		Irish & 43.66 & 18.72 & 15.86 & 41.93 & 17.16 & 15.25 \\
	\end{tabular}
	\caption{Extractive summarization baseline with modified LexRank. We report ROUGE F1 scores for both the validation and test splits.}
	\label{tab:monolingual}
\end{table}

One popular traditional algorithm for generating extractive summaries is LexRank~\cite{erkan-radev-2004-lexrank}.
We utilize a modified variant of LexRank that uses multilingual embeddings generated by sentence-transformers~\cite{reimers-gurevych-2019-sentence, reimers-gurevych-2020-making} to compute centrality.
Given the previously mentioned limitations of sentencizing input texts, we chunk the text based on existing paragraph separators (refer to ~\Cref{fig:outline}), and treat those segments as inputs to our baseline setup.
Notably, this method does not require any form of fine-tuning or language adoption and works as a zero-shot domain transferred extractive model, which makes it preferable over methods such as SummaRuNNer~\cite{nallapati-etal-2017-summarunner} or extractive BERT summarizers,\footnote{e.g., \texttt{bert-extractive-summarizer}} which require training on (automatically extracted) alignments.

\noindent To determine the output summary length, we calculate the average paragraph-level compression ratio on the language's training set, and then multiply this value with the reference document's number of paragraphs to obtain a target length. \\
For evaluation, we rely on ROUGE scores~\cite{lin-2004-rouge} with disabled stemming to conserve comparability between languages. We acknowledge that this is not a comprehensive measure and has distinctive shortcomings, but works fairly well at the paragraph level, as such units generally preserve both factual consistency and fluency.

\noindent Due to the paragraph-level consistency of generated summaries, this is a fairly strong baseline. Importantly, ROUGE scores remain consistent for languages between the validation and test set, although we do observe some languages with outlier performance: For Greek text, the model likely struggles with the representation of non-arabic subwords, but still performs decently well at the ROUGE-L level. Otherwise, Irish has unexpectedly high ROUGE scores, which we were unable to explain. This is especially surprising given the fact that the language is not even one officially supported by the multilingual embedding model used for this experiment.


\subsection{Cross-lingual Baselines}
As a baseline for XLS, we provide a simple two-step translate-then-summarize pipeline~\cite{wang2022survey}. To generate summaries on longer contexts, we utilize a model based on the Longformer Encoder Decoder (LED) architecture~\cite{beltagy2020longformer}, precisely a checkpoint previously fine-tuned on the English BillSum corpus~\cite{kornilova-eidelman-2019-billsum}. Translation from English to target languages is done with OPUS-MT~\cite{TiedemannThottingal:EAMT2020}. 
To deal with long documents exceeding the particular model's window size, we greedily chunk text if necessary.
To represent an upper limit of performance, we compare a translate-then-summarize setup from English to Spanish, which can be regarded as one of the language pairs with the highest MT performance, due to data availability and linguistic similarity of the source and target language.

\noindent As baselines, we provide translations of the English gold summaries into the target language (again with the Opus MT model), as well as a translation of the extractive LexRank summary from the previous experiment.
Results seen in \Cref{tab:crosslingual} are surprising: While the abstractive model seems to improve over the purely Spanish-based LexRank summary (LexRank-ES) by a significant margin, it turns out that translating the English LexRank baseline drastically \emph{improves} results in terms of ROUGE scores. We assume that this is related to truncation and re-phrasing happening during the translation step.

\begin{table}[t]
	\setlength{\tabcolsep}{1.5pt}
	\centering
	\hspace*{-0.7em}
	\begin{tabular}{l|ccc||ccc}
		&  \multicolumn{3}{c||}{\textbf{Validation} } & \multicolumn{3}{c}{\textbf{Test}} \\
		& \textbf{R-1} & \textbf{R-2} & \textbf{R-L} & \textbf{R-1} & \textbf{R-2} & \textbf{R-L} \\ 
		\hline
		LED & 31.67 & 13.00 & 16.17 & 31.14 & 13.01 & 16.20 \\
		LexRank-EN & 39.42 & 20.03 & 18.53 & 39.44 & 20.02 & 18.73 \\
		\hline
		LexRank-ES & 27.04 & 16.43 & 14.75 & 28.34 & 17.12 & 15.23 \\
		Oracle & 52.84 & 39.79 & 43.87 & 54.55 & 41.01 & 45.06 \\
	\end{tabular}
	\caption{Cross-lingual summarization setup for English-Spanish. We report ROUGE F1 scores for both the validation and test splits on the Spanish subset.}
	\label{tab:crosslingual}
\end{table}

\subsection{Open Problems}
The most obvious problem for this dataset is the extreme length, and also length disparity between documents. This is especially apparent when comparing the length to average samples in CNN/DailyMail~\cite{DBLP:conf/nips/HermannKGEKSB15}, where the mean article length is about 16 times shorter; this makes content selection significantly more challenging.

\noindent Secondly, incorporating hierarchical information about the reference text could greatly improve context relevance in such extensive settings.
However, this is not only restricted to the reference document, but could also be considered for the (hierarchical) construction of long-form summary texts.
Given that previous datasets do not come with such long output samples, this has to our knowledge not been previously tackled in the literature.

\noindent Ultimately, the question of equal coverage for lesser-resourced languages is also not fully answered. While we attempt to treat languages in our dataset equally, this comes with its particular set of challenges and performance hits in highly available languages.



\section{Conclusion and Future Work}

Throughout this work, we have detailed the creation of a new multilingual corpus for text summarization, based on legal acts from the European Union.
We further provided a more detailed analysis of the underlying data and sample quality and hypothesized potential applications to open problems in the communtiy, such as long-form summarization or cross-lingual application scenarios.
Our dataset is publicly available on the web, and comes with a set of monolingual extractive baselines that provide suitable reference points for any future work in this direction.\\
In particular, we intend to focus on exploiting the structure of summaries for a more guided generation of output texts.
Especially for extremely long legal texts, template structures could be utilized. On a more general level, we expect that progress in long-form models is required to achieve remotely sensible results on extreme-length generative tasks. Alternative approaches in the meantime could include aspect-driven methods for building summaries in an iterative fashion.

\noindent Finally, on top of the static snapshot presented in this work, we are also working towards a continually updated data repository of this resource, which would then include newly added texts (or summaries) for EU texts.

%
\section{Limitations}
While our work considers comparatively high-quality data samples, there still remain some assumptions about the underlying text sources, which lead to some of the following limitations:

\begin{enumerate}
	\item Documents themselves (both sources and summaries) may link to external articles or related regulations for further information. Some of the linked documents might indeed contain relevant contextual information, but are as such not considered in our version.
	\item On a similar note, we mentioned that some summaries aggregate content from \emph{several legal acts}, as outlined in \Cref{sec:filtering}; this is not considered in full at the current stage and might cause limitations.
	\item Legal acts may exist in several iterations, drafted up at different points in time. To the best of our knowledge, we extracted the most recent version and its associated summary.
	\item For evaluation of generated summarization quality, we provide $n$-gram-based ROUGE scores, which have previously been argued to poorly reflect particular aspects, e.g., factual consistency or fluency. Given this, baseline performance should be taken in clear context for future work.
\end{enumerate}

\section*{Broader Impact \& Ethical Issues}
With the release of our data as a public resource, we want to touch on the potential ethical implications of this release: As our data is already available (though much more inaccessible) through the EUR-Lex platform, we do not see any ethical concerns in a repurposed and bundled release of this dataset from such a standpoint.
To our knowledge, human-written reference and summary texts, as well as the accompanying translations into the European languages, have undergone review within instances of the European Union, leading to no clear concerns in data quality, especially with respect to potential privacy violations or harmful text content.\\
However, we acknowledge that this resource reinforces a certain availability bias towards European languages, which needs to be acknowledged by follow-up work.
On the other hand, we believe that the release of resources including underrepresented languages, for example, Irish and Maltese, outweighs this concern and fosters future research in a more multilingual way.

\section*{Acknowledgements}
We want to especially thank Reviewer 1 for their constructive feedback and suggestion of further experiments in their review.

\bibliography{anthology,custom}
\bibliographystyle{acl_natbib}

\appendix
\section{Language-specific Distributions}
\begin{figure*}[ht]
	\centering
	\includegraphics[width=1.0\textwidth]{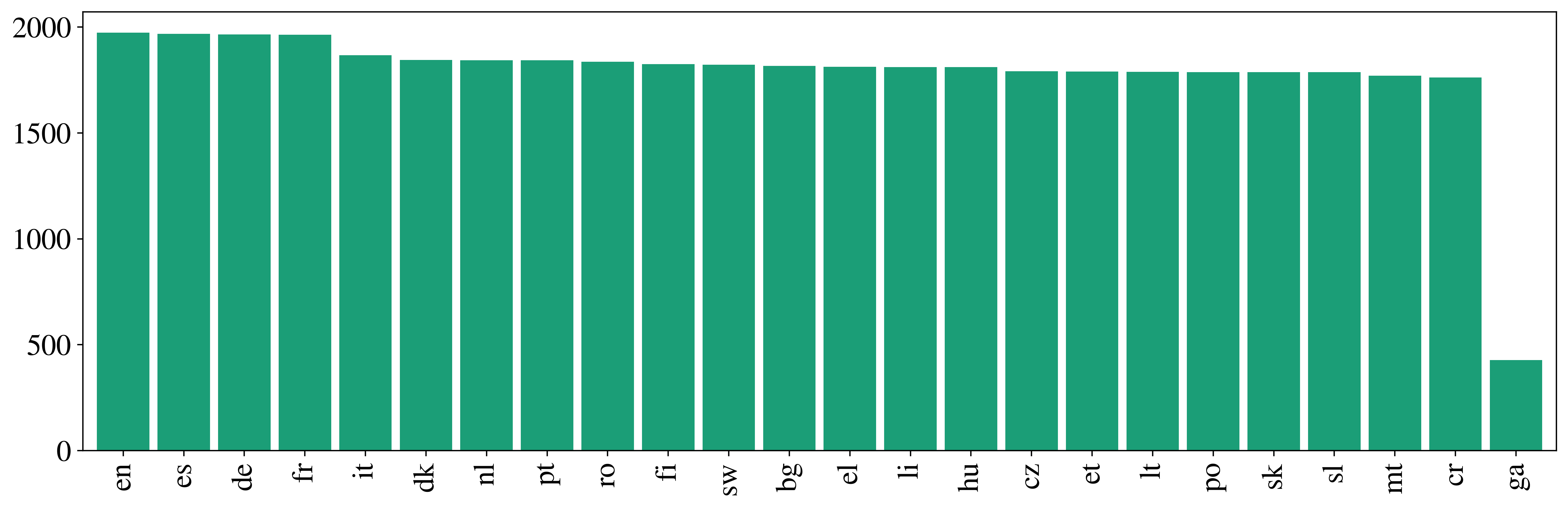}
	\caption{The number of all crawled document/summary pairs across the 24 official EU languages \emph{before} filtering. Irish exhibits a greatly limited availability due to its recent addition as an official language.}
	\label{fig:langs}
\end{figure*}

\begin{figure}[t]
	\centering
	\includegraphics[width=0.45\textwidth]{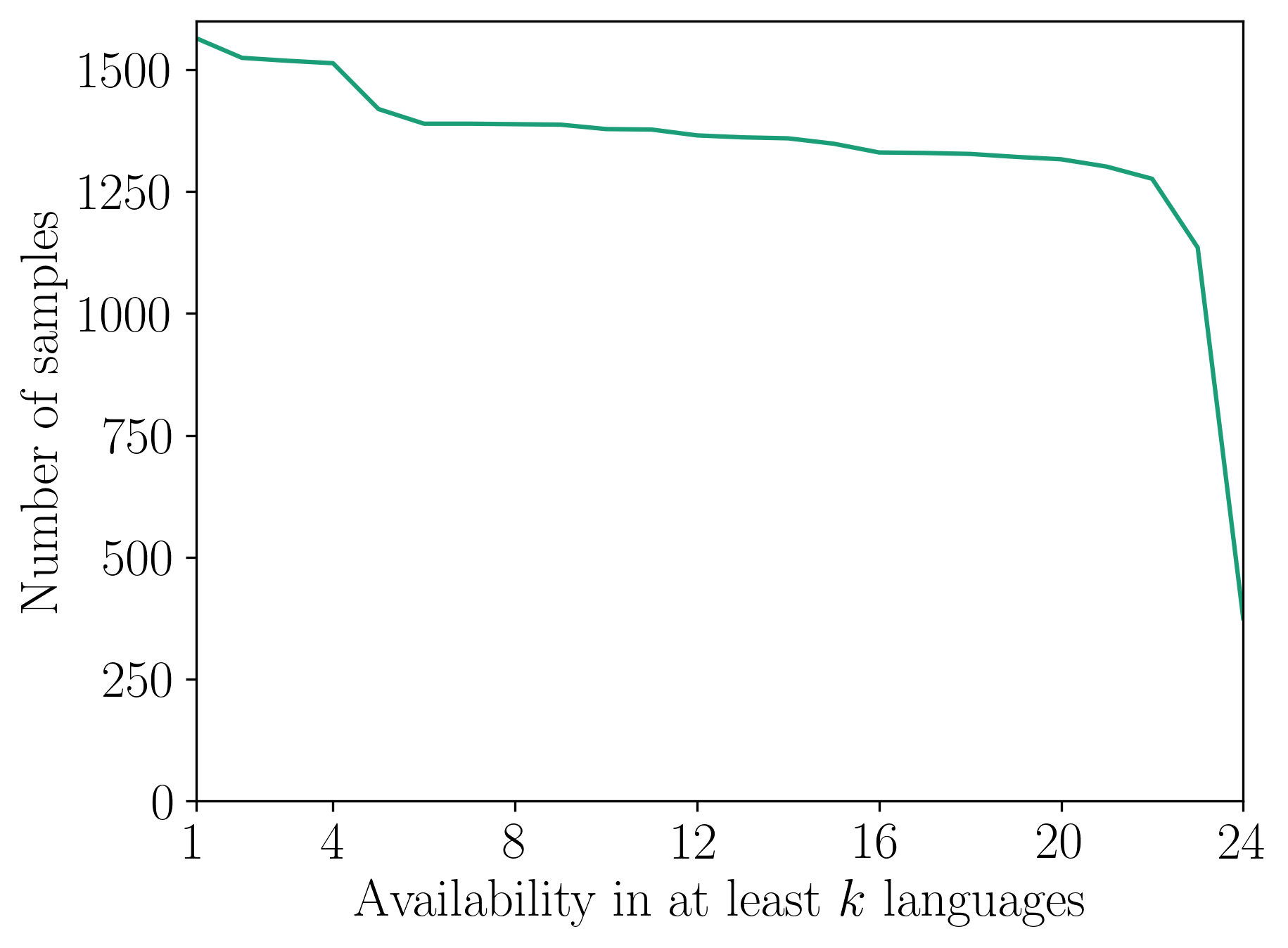}
	\caption{Celex IDs present in at least $k$ languages.}
	\label{fig:samples}
\end{figure}

\Cref{fig:langs} displays available articles before filtering. Notably, the documents need not be subsets of one another, meaning the French document IDs might differ from English ones.
\Cref{tab:langs} further compares the availability of language-specific articles before and after filtering, to provide an insight into the number of removed documents.
The same table also provides a more concise overview of supported languages in popular frameworks, as well as an extension of statistics reported in \Cref{tab:en_stats} for the language-specific training sets.
To illustrate cross-lingual presence of Celex IDs, we plot the inverse availability distribution (sample is available in \emph{at least $k$ languages}) in \Cref{fig:samples}. Around 84\% of the samples are available in 20 or more languages.

%
%

\begin{table*}
	\setlength{\tabcolsep}{2.5pt}
	\centering
	\begin{tabular}{l|cc|ccc|c|c|c|cc}
			 & \multicolumn{2}{c|}{\textbf{No.~articles} } & \multicolumn{3}{c|}{\textbf{Availability}} & \multicolumn{2}{c|}{\textbf{Article token length}} & \textbf{Comp.} & \multicolumn{2}{c}{\textbf{$n$-gram novelty}} \\
		\textbf{Language} & \textbf{before} & \textbf{after} & \textbf{S-T} & \textbf{spacy} & \textbf{nltk} & \multicolumn{1}{c}{\textbf{Reference}} & \textbf{Summary} & \textbf{ratio} & \multicolumn{1}{c}{\textbf{$1$-gram}} & \textbf{$2$-gram}\\ 
		\hline
		English (en) & 1{,}974 & 1{,}504 & \cmark & \cmark & \cmark & 12206 $\pm$ 42429 & 799 $\pm$ 349 & 16 $\pm$ 62 & 44.10 & 65.97 \\
		French (fr) & 1{,}969 & 1{,}505 & \cmark & \cmark & \cmark & 13192 $\pm$ 43950 & 892 $\pm$ 395 & 16 $\pm$ 63 & 45.07 & 64.13 \\
		German (de) & 1{,}966 & 1{,}490 & \cmark & \cmark & \cmark & 11144 $\pm$ 41061 & 748 $\pm$ 330 & 16 $\pm$ 68 & 44.85 & 66.95 \\
		Spanish (es) & 1{,}964 & 1{,}487 & \cmark & \cmark & \cmark & 13581 $\pm$ 44574 & 932 $\pm$ 420 & 15 $\pm$ 57 & 44.76 & 61.51 \\
		Italian (it) & 1{,}867 & 1{,}403 & \cmark & \cmark & \cmark & 13152 $\pm$ 44641 & 845 $\pm$ 370 & 16 $\pm$ 67 & 44.77 & 67.00 \\
		Portuguese (pt) & 1{,}845 & 1{,}376 & \cmark & \cmark & \cmark & 12629 $\pm$ 29921 & 896 $\pm$ 391 & 14 $\pm$ 38 & 43.84 & 64.00 \\
		Dutch (nl) & 1{,}844 & 1{,}376 & \cmark & \cmark & \cmark & 13233 $\pm$ 44638 & 834 $\pm$ 362 & 17 $\pm$ 69 & 44.41 & 65.86 \\
		Danish (da) & 1{,}843 & 1{,}377 & \cmark & \cmark & \cmark & 11947 $\pm$ 43155 & 717 $\pm$ 308 & 18 $\pm$ 71 & 46.96 & 68.27 \\
		Greek (el) & 1{,}837 & 1{,}366 & \cmark & \cmark & \cmark & 13609 $\pm$ 45411 & 863 $\pm$ 369 & 17 $\pm$ 64 & 44.86 & 66.70 \\
		Finnish (fi) & 1{,}825 & 1{,}366 & \cmark & \cmark & \cmark &\ \ 9792 $\pm$ 41021 & 575 $\pm$ 247 & 18 $\pm$ 93 & 53.41 & 77.26 \\
		Swedish (sv) & 1{,}822 & 1{,}362 & \cmark & \cmark & \cmark & 10796 $\pm$ 26923 & 718 $\pm$ 305 & 15 $\pm$ 40 & 46.74 & 69.62 \\
		Romanian (ro) &1{,}817 & 1{,}353 & \cmark & \cmark & \xmark & 13646 $\pm$ 45644 & 826 $\pm$ 356 & 17 $\pm$ 67 & 45.42 & 67.80 \\
		Hungarian (hu) & 1{,}813 & 1{,}336 & \cmark & \xmark & \xmark & 12230 $\pm$ 46764 & 702 $\pm$ 298 & 19 $\pm$ 84 & 53.23 & 75.68 \\
		Czech (cs) & 1{,}812 & 1{,}359 & \cmark & \textbf{?} & \cmark & 12469 $\pm$ 46640 & 715 $\pm$ 307 & 18 $\pm$ 77 & 46.75 & 71.89 \\
		Polish (pl) & 1{,}811 & 1{,}353 & \cmark & \cmark & \cmark & 11560 $\pm$ 33296 & 739 $\pm$ 324 & 16 $\pm$ 48 & 46.69 & 71.01 \\
		Bulgarian (bg) & 1{,}792 & 1{,}332 & \cmark & \xmark & \xmark & 13397 $\pm$ 45578 & 819 $\pm$ 350 & 17 $\pm$ 69 & 47.00 & 68.44 \\
		Latvian (lv) & 1{,}790 & 1{,}334 & \cmark & \textbf{?} & \xmark & 11841 $\pm$ 46552 & 670 $\pm$ 289 & 19 $\pm$ 83 & 50.23 & 74.55 \\
		Slovene (sl) & 1{,}789 & 1{,}332 & \cmark & \xmark & \cmark & 11357 $\pm$ 32842 & 712 $\pm$ 305 & 16 $\pm$ 48 & 47.28 & 71.57 \\
		Estonian (et) & 1{,}788 & 1{,}332 & \cmark & \xmark & \cmark & 10778 $\pm$ 45157 & 581 $\pm$ 249 & 20 $\pm$ 94 & 52.20 & 77.46 \\
		Lithuanian (lt) & 1{,}788 & 1{,}335 & \cmark & \textbf{?} & \cmark & 11943 $\pm$ 46673 & 669 $\pm$ 290 & 19 $\pm$ 88 & 47.79 & 74.00 \\
		Slovak (sk) & 1{,}788 & 1{,}325 & \cmark & \textbf{?} & \xmark & 11600 $\pm$ 32968 & 729 $\pm$ 319 & 16 $\pm$ 47 & 48.20 & 73.42 \\
		Maltese (mt) & 1{,}770 & 1{,}315 & \xmark & \xmark & \xmark & 12711 $\pm$ 48156 & 685 $\pm$ 299 & 20 $\pm$ 85 & 54.77 & 81.43 \\
		Croatian (hr) & 1{,}762 & 1{,}278 & \cmark & \textbf{?} & \xmark & 10051 $\pm$ 19390 & 712 $\pm$ 307 & 14 $\pm$ 28 & 48.62 & 72.61 \\
		Irish (ga) & 427 & 391 & \xmark & \textbf{?} & \xmark & 28152 $\pm$ 63360 & 948 $\pm$ 385 & 46 $\pm$ 137 & 45.89 & 70.38 \\
	\end{tabular}
	\caption{Supplementary statistics of the EUR-Lex-Sum corpus across languages. We list the total number of available articles (before and after filtering), and whether a particular language is supported by \texttt{sentence-transformers} multilingual models (``\emph{S-T}''), or has available language-specific models in \texttt{spaCy}~\cite{honnibal-etal-2020-spacy} or \texttt{nltk}~\cite{bird-etal-2009-natural}, respectively. ``\textbf{?}'' indicates potential support through general-purpose models with uncertain segmentation quality. We also provide abriged statistics along the lines of \Cref{fig:stats} and \Cref{tab:en_stats} for the training partition of all languages.}
	\label{tab:langs}
\end{table*}

\section{Implementation Details for Baselines}
\sloppy
For extractive monolingual models, we use the checkpoint ``paraphrase-multilingual-mpnet-base-v2''\footnote{model configuration: \url{https://huggingface.co/sentence-transformers/paraphrase-multilingual-mpnet-base-v2/blob/main/config.json}, last accessed: 2022-06-23} without any further fine-tuning, using version 2.1.0 of \texttt{sentence-transformers}.
We do use a slightly modified version of their LexRank implementation to avoid a bug preventing the power method from converging due to ``negative likelihoods''. This can be fixed by normalizing similarity scores to strictly positive values.\\
The abstractive LEDBill model\footnote{\url{https://huggingface.co/d0r1h/LEDBill}, last accessed: 2022-06-23} was used through the \texttt{pipeline} feature available in Huggingface Transformers~\cite{wolf-etal-2020-transformers}, version 4.18.
We use greedy decoding for text generation and chunk text into blocks of approximately 4096 tokens, where we then concatenate the output summaries of consecutive sections.
A similar setup was used for translation, where we use the Opus MT models for respective language pairs (\texttt{HelsinkiNLP/opus-mt-<src>-<tgt>}) \cite{TiedemannThottingal:EAMT2020}, although the context size used for chunking is 500 subword tokens (to account for model-specific padding). We refer to the model card for configuration hyperparameters of LEDBill\footnote{\url{https://huggingface.co/d0r1h/LEDBill/blob/main/config.json}, last accessed: 2022-06-23} and Opus MT\footnote{\url{https://huggingface.co/Helsinki-NLP/opus-mt-en-es/blob/main/config.json}, last accessed: 2022-06-23}.
Parameter counts for all three neural models can be found in \Cref{tab:counts}.

\noindent For the computation of ROUGE scores, we utilize the implementation by Google Research\footnote{\url{https://pypi.org/project/rouge-score/}; version 0.0.4, last accessed: 2022-06-23}, with stemming disabled.\\
For GPU inference, we use a machine with a single Nvidia Titan RTX with 24 GB GPU VRAM and 64 GB RAM.
Obtaining results for the LexRank baselines on both the test and validation set takes less than 2.5 minutes on average for the validation and test samples (375 total generated summaries). In comparison, the generation with LEDBill takes approximately 12 hours per 375 validation/test samples.
Computationally speaking, translations lie somewhere in between the previous settings, taking around 20 minutes to compute.

\begin{table}[h]
	\centering
	\begin{tabular}{l|c}
		\textbf{Model} & \textbf{Parameters} \\
		\hline
		S-T & 278MM \\
		LEDBill & 162MM \\
		Opus MT & 78MM
	\end{tabular}
	\caption{Approximate parameter count for utilized neural systems. ``S-T'' represents the sentence-transformers model used for computing sentence embbeddings in the modified LexRank baseline.}
	\label{tab:counts}
\end{table}


\end{document}